\pdfoutput=1

\documentclass[11pt]{article}

\usepackage{acl}

\usepackage{times}
\usepackage{latexsym}
\usepackage{graphicx}

\usepackage[T1]{fontenc}

\usepackage[utf8]{inputenc}

\usepackage{microtype}
\usepackage{bm}
\usepackage{amsmath}
\usepackage{multirow}
\usepackage{booktabs}
\usepackage{subfigure}
\usepackage{CJKutf8}
\usepackage{pifont}

\usepackage{algorithm}
\usepackage{algpseudocode}

\title{BigTranslate: Augmenting Large Language Models with Multilingual Translation Capability over 100 Languages}

\author{Wen Yang\textsuperscript{1,2}, Chong Li\textsuperscript{1,2}, Jiajun Zhang\textsuperscript{1,2,3}\thanks{\ \ Corresponding author.}, and Chengqing Zong\textsuperscript{1,2} \\
\textsuperscript{1} Institute of Automation, Chinese Academy of Sciences \\
\textsuperscript{2} School of Artificial Intelligence, University of Chinese Academy of Sciences \\
\textsuperscript{3} Wuhan AI Research\\
\texttt{$\left\{\right.$yangwen2023, lichong2021$\left .\right\}$@ia.ac.cn}\\
\texttt{$\left\{\right.$jjzhang, cqzong$\left .\right\}$@nlpr.ia.ac.cn}
}

\begin{document}
\begin{CJK*}{UTF8}{gbsn}
\maketitle
\begin{abstract}
Large language models (LLMs) demonstrate promising translation performance among various natural languages. However, many LLMs especially the open-sourced ones, such as BLOOM \citep{scao2023bloom} and LLaMA \cite{touvron2023llama}, are English-dominant and support only dozens of natural languages, making the potential of LLMs on language translation less explored. In this work, we present BigTranslate which adapts LLaMA that covers only 20 languages and enhances it with multilingual translation capability on more than 100 languages. BigTranslate is built upon LLaMA-13B and it is optimized in three steps. First, we continue training LLaMA with massive Chinese monolingual data. Second, we continue training the model with a large-scale parallel dataset that covers 102 natural languages. Third, we instruct-tune the foundation model with multilingual translation instructions, leading to our BigTranslate model. The preliminary experiments on multilingual translation show that BigTranslate performs comparably with 
ChatGPT and Google Translate in many languages and even outperforms ChatGPT in 8 language pairs. We release the BigTranslate model\footnote{https://github.com/ZNLP/BigTranslate} and hope it can advance the research progress.

\end{abstract}

\section{Introduction}
Large language models (LLMs), such as ChatGPT \citep{openai2022chatgpt} and PaLM 2 \cite{anil2023palm}, demonstrate impressive translation capabilities among various natural languages. For example, several recent studies indicate that ChatGPT is a good translator in many scenarios (e.g., spoken language translation, document translation, and multilingual translation), and it can even outperform SOTA translation engines in some specific scenarios \cite{bawden2023investigating,hendy2023good,jiao2023chatgpt,wang2023document,zhu2023multilingual}. LLMs are also preferred as a translator for their interactive usage.

However, most of the existing LLMs are English-dominant and the popular LLMs support only several or dozens of natural languages. For example, GLM \citep{du-etal-2022-glm, zeng2022glm} just supports English and Chinese. BLOOM \citep{scao2023bloom} covers 46 languages while LLaMA \citep{touvron2023llama} only supports 20 languages. It is well-known that there are over 7000 natural languages in the world and existing LLMs cover only a very small fraction of the languages. Obviously, a large population in the world cannot be benefited from the multilingual capability of LLMs.

In order to equip LLMs with much more multilingual ability, we introduce BigTranslate that adapts LLaMA making it capable of translating over 100 natural languages. Instead of optimizing the foundation model with self-supervised learning on massive monolingual data over multiple languages, we mainly employ the bitexts which can transfer the knowledge from high-resource languages to low-resource ones through semantic mapping between parallel sentences.

Specifically, we build BigTranslate based on the 13B version of LLaMA which is proven to be comparable with GPT-3 in many natural language processing tasks. The BigTranslate is constructed in three steps. In the first step, we utilize massive Chinese texts to continue training LLaMA, resulting in a strong model which well supports Chinese. In the second step, a large-scale parallel dataset covering 102 natural languages is employed to continue training the LLMs in a curriculum learning manner and we obtain a multilingual foundation model that has the potential to perform translating among more than 100 natural languages. In the third step, instruction tuning is applied to optimize the multilingual foundation model with rich translation instructions. Finally, we get our BigTranslate model.

To verify the effectiveness of our BigTranslate model, we conduct preliminary multilingual translation experiments on all 102 languages. We compare BigTranslate with both Google Translate and ChatGPT. Since the automatic evaluation metric BLEU is usually criticized for the poor correlation with human judgments in machine translation quality, we further employ GPT-4 \citep{openai2023gpt4} which shows a high correlation with human \cite{liu2023geval} as the evaluator and we design well-defined prompts to make GPT-4 act like a human evaluator. The experiments show that BigTranslate performs comparably with Google and ChatGPT in many languages, and even outperforms ChatGPT in 8 language pairs.

\section{Related Work}
\subsection{Large Language Models}
Since the advent of GPT-3 \cite{brown2020language} by OpenAI in 2020, large language models that employ Transformer as the backbone and contain tens or hundreds of billions of parameters, such as PaLM \cite{chowdhery2022palm}, OPT \cite{zhang2022opt}, BLOOM \cite{scao2023bloom}, Chinchilla \cite{hoffmann2022training}, Galactica \cite{taylor2022galactica}, GLM \cite{du-etal-2022-glm,zeng2022glm}, and LLaMA \cite{touvron2023llama}, are constantly emerging. Among these LLMs, ChatGPT is a big milestone that demonstrates that the foundation large language model exhibits emergent and general abilities when the model is large enough and equipped with instruction tuning. 

Many recent studies investigate the ability of ChatGPT, GPT-4, and other LLMs in traditional natural language processing tasks, including machine translation \cite{bawden2023investigating,hendy2023good,jiao2023chatgpt,wang2023document,zhu2023multilingual}. \cite{jiao2023chatgpt} reports that ChatGPT and GPT-4 are good translators, especially for high-resource languages and spoken language scenarios. \cite{wang2023document} demonstrates that ChatGPT and GPT-4 perform quite well in document translation with the help of long context modeling. \cite{bawden2023investigating} and \cite{zhu2023multilingual} show that LLMs like ChatGPT and BLOOM are also multilingual translators and even outperform SOTA online translation engines. However, most of the existing LLMs are English-dominant and cover up to only dozens of languages. In this work, we address this challenge and present BigTranslate which can support more than 100 languages.

\subsection{Multilingual Neural Machine Translation}
Multilingual neural machine translation aims at translating multiple languages with a single shared model \cite{johnson2017google}. Most studies focus on the unbalanced problem in multilingual translation. For example, some works investigate how to design shared and language-dependent model parameters in a multilingual translation framework \cite{Wang20118three,wang2019compact,lin2021learning,Xie2021importance,wang2022parameter}. Several works explore how to train the multilingual translation model more effectively and efficiently when the training data are quite unbalanced across languages \cite{zhou2021distributionally,huang2022unifying}. Few studies pursue the potential of a multilingual translation model on handling more than 100 languages. For example, NLLB \cite{costa2022no} proposed by Meta aims to build a multilingual translation model that could translate as many languages as possible (currently covering more than 200 languages). However, this kind of model can only perform translating. In this work, we pursue constructing a multilingual translator by adapting an LLM while maintaining its generic ability.

\section{BigTranslate Construction}

\subsection{LLaMA as Foundation Model}
Considering its impressive performance on most English benchmarks after pre-training on 1.4T tokens \citep{touvron2023llama}, LLaMA is adopted as our foundation model. 
Specifically, the BigTranslate is initialized from the LLaMA-13B model to reduce the computational cost and continues to train on massive Chinese and parallel corpus.

\subsection{Augmenting Foundation Model with Chinese}
\label{section_3.2}
LLaMA shows a poor Chinese language understanding and generation performance due to the lack of a sufficient Chinese pre-training corpus \citep{cui2023efficient}, although its performance in English is comparable to or even better than GPT-3.
Moreover, \citet{shah2023geometry} found that Chinese exhibits a significantly low cross-lingual similarity with other languages, including English and German, which means that the inferior processing ability in Chinese text will hinder our foundation model towards a better multilingual translation model.
Thus it is indispensable to augment the ability of our foundation model with additional Chinese vocabulary and pre-training corpus. By doing so, we expect the final model to be capable of multilingual translation centered on both English and Chinese.

To achieve this, we first append the original vocabulary with 6,223 Chinese tokens, most of which are Chinese characters, generated from SentencePiece \citep{kudo-richardson-2018-sentencepiece} using the byte-pair encoding (BPE) algorithm \citep{sennrich-etal-2016-neural} on Chinese text.
Then, a large-scale Chinese pre-training corpus, including CLUE \citep{xu2019nlpcorpus, xu-etal-2020-clue}, Chinese News, and Chinese question-answering dataset, is adopted to pre-train the model resulting in Chinese LLaMA. 

\subsection{Augmenting Foundation Model with 102 languages}
The purpose of this study is to equip a large language model with Chinese-centric multilingual ability. Despite intensive training on massive Chinese monolingual data, the Chinese LLaMA model primarily exhibits proficiency in Chinese language processing, but lacks adequate multilingual capabilities. Furthermore, the continuous training on Chinese monolingual data has the potential to diminish the model's performance in the 20 languages originally supported by LLaMA. To address these limitations, we further refine the foundational model by incorporating a substantial parallel dataset encompassing 102 languages. This second training stage aims to enhance the model's ability to facilitate multilingual translation tasks, enabling it to support a broader range of languages.

Large language model pre-training is typically conducted using monolingual data, focusing on autoregressive training in the language to enhance the model's capabilities. In contrast, multilingual large language model pre-training involves training on multilingual parallel data, enabling the language model to learn across different languages. The primary challenge in multilingual large language model pre-training lies in achieving a balance in learning between high-resource and low-resource languages. This balance ensures that the model acquires proficiency in both high-resource and low-resource languages during training.

The issue of maintaining balanced learning between high-resource and low-resource languages has been a long-time topic of concern in multilingual learning. Despite the notable progress made in the development of large language models, this problem is still not well solved as the model capacity increases. To address this challenge encountered during the pre-training phase, our work proposes an incremental data sampling strategy. By employing the strategy, the model is trained on a harmonized multilingual parallel corpus at each stage, mitigating the concern of unbalanced language proficiency.

\subsubsection{Large-scale Parallel Dataset Construction}
In order to enhance the language capabilities of the Chinese LLaMA model to support 102 languages, we constructed a comprehensive parallel corpus dataset consisting of 102 languages. This dataset was employed to continue training the foundational model. The compilation of this dataset drew upon multiple sources, including widely available public parallel corpus datasets and household datasets. The public datasets utilized in our study contain IWSLT, WMT, CCMT, and OPUS-100 \citep{zhang-etal-2020-improving}, forming the initial corpus of our dataset.

After obtaining the initial corpus, a key consideration is to balance the two-way translation in language pairs. For example, the data size of Chinese-to-English translation should be similar to English-to-Chinese. To achieve this balance, we utilize data augmentation to enrich the corpus if necessary. The data augmentation follows the subsequent strategy. In cases where the number of parallel corpus falls below 1 million, we flip the entire corpus to create the corpus for the opposite translation direction. In contrast, for corpora with more than 1 million instances, we randomly flip half the amount of corpus to generate the corresponding corpus. After data augmenting, the initial corpus of 142 translation directions is substantially enriched, expanding to a significantly larger corpus of 242 translation directions.

\begin{figure*}[htbp]
    \centerline{
    \includegraphics[width=16cm]{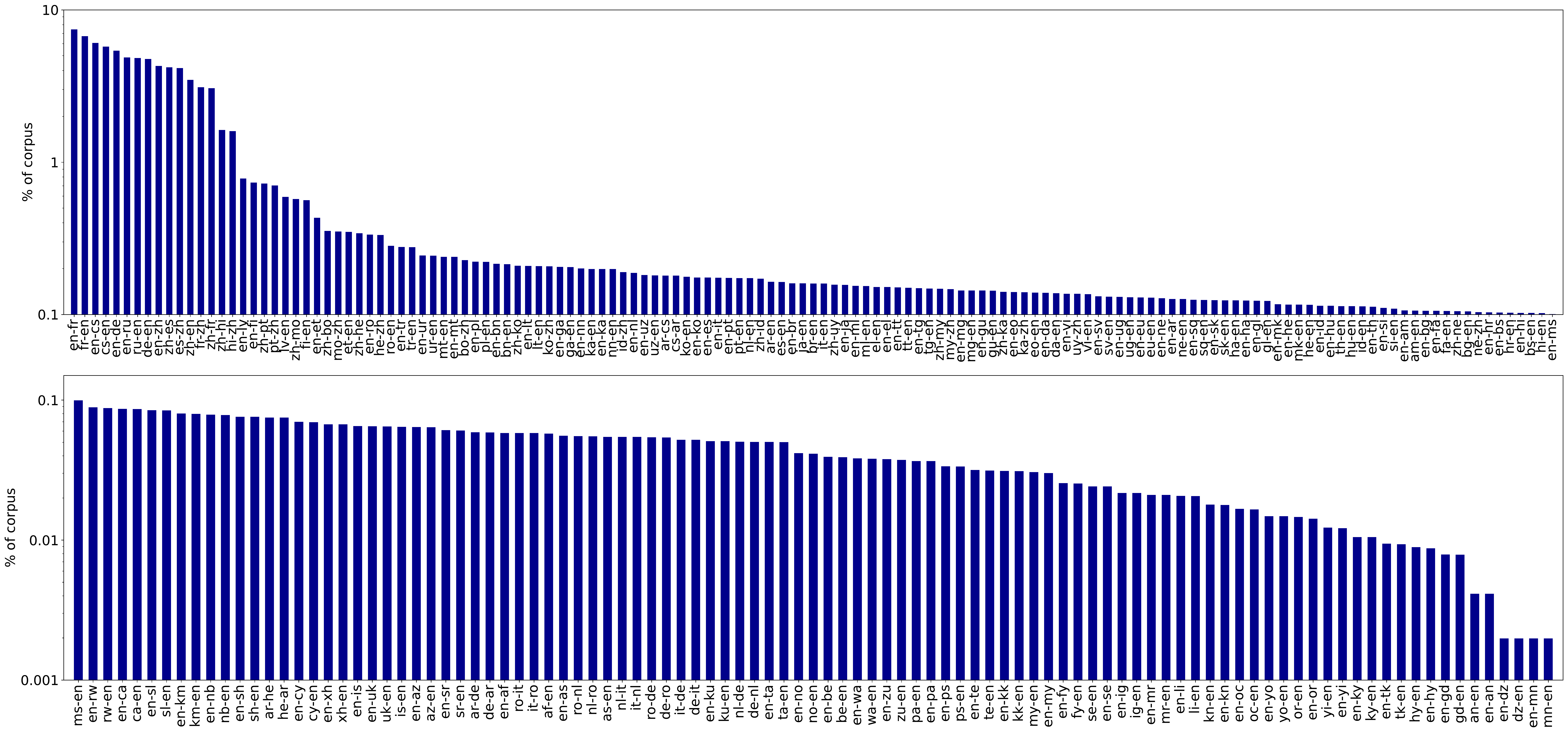}
    }
    \caption{The language-pairs distribution of multilingual corpus. All the datasets consist of about 300 million sentence pairs.}
    \label{fig_distribution}
\end{figure*}

To effectively illustrate the distribution of the corpus, we present a visual representation of the language-pair distribution within the multilingual datasets as figure \ref{fig_distribution}. The matter pertaining to the imbalance between high-resource and low-resource language pairs continues to be a prominent concern within the current corpus.

\subsubsection{Tokenizer}

LLaMA tokenizes data with the byte-level byte-pair (BBPE) encoding algorithm, implemented by SentencePiece. LLaMA's original vocabulary includes 32,000 tokens, primarily comprising English and Latin tokens. To augment LLaMA's proficiency in handling Chinese text, we expand its vocabulary by incorporating additional 6,223 Chinese tokens as section \ref{section_3.2} introduced. Furthermore, to boost LLaMA's comprehension across multilingual parallel datasets, we further extend the vocabulary to 53,613 tokens, most of the added tokens are trained on texts spanning 102 languages.

To alleviate the issue of vocabulary skew resulting from imbalances in instances across the corpus, we implement a strategy that involved selecting a subset of the large-scale parallel dataset for vocabulary training. Specifically, we randomly sample \textit{max\_num} instances from each language to be included in the vocabulary training. This approach ensures that the common words shared among 102 languages are adequately represented, serving as a crucial prerequisite for model understanding multi-languages. Concretely, we set \textit{max\_num} to 1,000,000.

\subsubsection{Incremental Multilingual Pre-training}

Prior to data sampling, we employ multilingual vocabulary to segment the entire multilingual parallel corpus. Subsequently, we construct training samples by concatenating the same language sentence pairs. Each sample is comprised of multiple parallel sentence pairs and has a fixed length of 1,024 tokens. This approach ensured the formation of coherent and consistent training samples for subsequent model training.

To mitigate the issue of the model disproportionately focusing on learning high-resource corpus during the training phase, which could potentially hinder the learning of low-resource languages, we draw inspiration from curriculum learning \citep{bengio2009curriculum} to propose an incremental approach for multilingual LLMs pre-training.

In this incremental pre-training method, we gradually expose the model to language pairs in a curriculum-like manner. Initially, the model is exposed to high-resource language pairs, allowing it to establish a solid foundation in those languages. Subsequently, we progressively introduce low-resource language pairs, enabling the model to gradually expand its knowledge and proficiency in these languages.

\begin{figure*}[htbp]
    \centerline{
    \includegraphics[width=16cm]{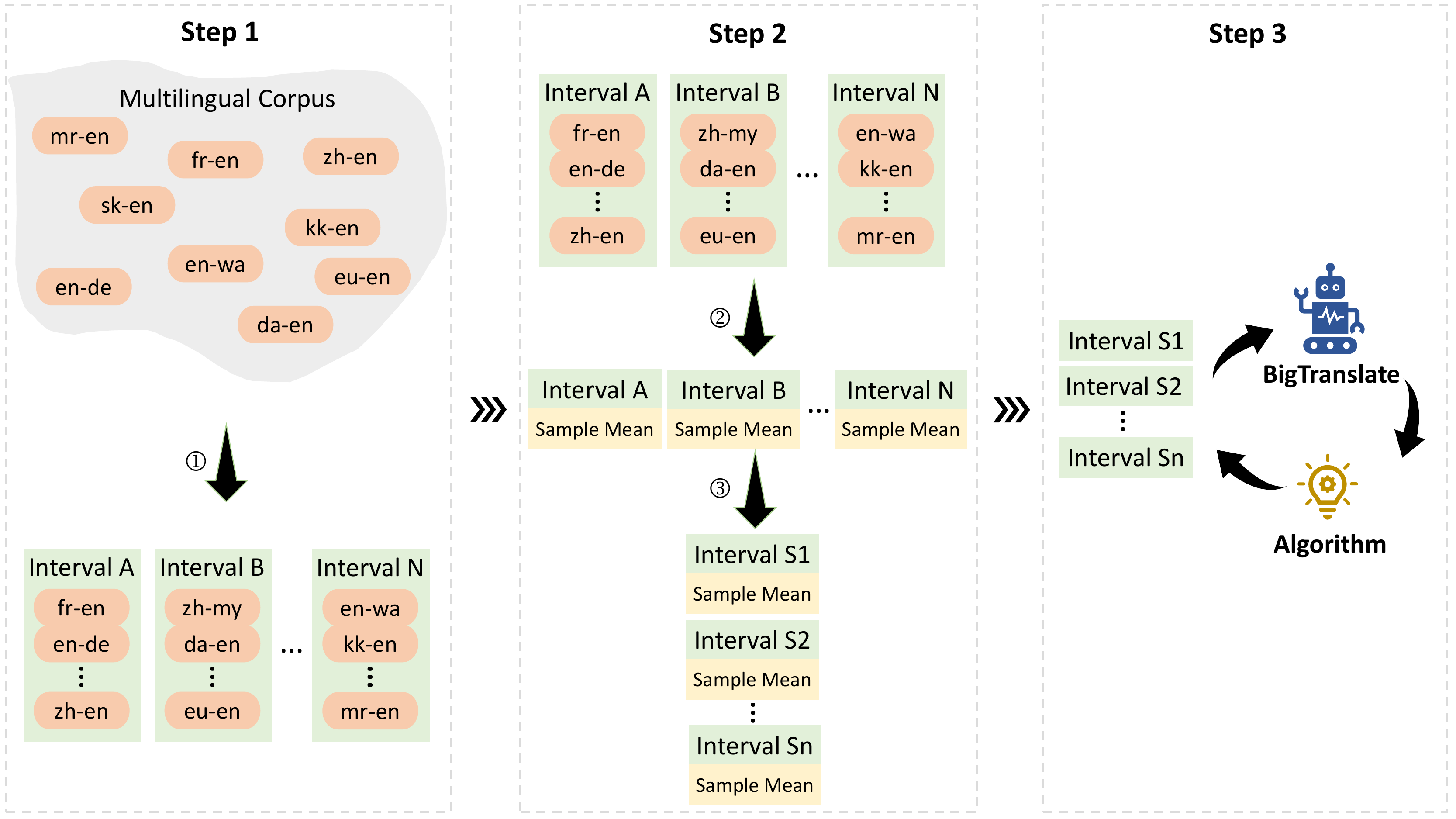}
    }
    \caption{The outline of three-step incremental multilingual pre-training approach. \ding{172} represents dividing multilingual language pairs into different intervals, \ding{173} denotes calculating sample means for all language pairs within each sample interval, \ding{174} represents sorting the intervals in descending order based on sample mean values. The algorithm in step 3 is detailed in \textbf{Algorithm \ref{alg:Framwork}} for incremental pre-training.}
    \label{fig_outline}
\end{figure*}

Specifically, we follow a three-step approach in our incremental pre-training method, as shown in Figure \ref{fig_outline}. Firstly, we set the sample interval size and divide language pairs into distinct intervals based on the number of instances for each language pair. Secondly, we calculate the sample mean for all language pairs in each interval. Thirdly, we dynamically measure the moment of adding the language-pair samples next interval according to the sample means in the previous sample interval. In the following part, we detail the three steps.

In the first step, we set sample interval size, denoted as \textit{S}. We hypothesize that if the interval size \textit{S} is below a specific threshold, we can consider that all language pairs in this interval have average resources. Consequently, training the multilingual model on this interval enables the model to acquire the average proficiency in each language.

In our preliminary experiments, it is observed that an excessively large language interval results in an unbalanced learning distribution among language pairs within the interval. Conversely, setting the interval too small leads to the model focusing on only a few or even just one language within the interval. Additionally, we noticed that high-resource and low-resource languages exhibited different sensitivities to interval size. To minimize the learning discrepancy between language pairs, we determine the interval size for language pairs based on their respective sample sizes. For language pairs with a sample size exceeding 10,000, we set the interval size \textit{S} to 10,000. Conversely, for language pairs with a sample size below 10,000, we opted for a sample interval \textit{S} of 5,000.

For example, the sample size of En-Ro (\textit{English-to-Romanian}) pair is 80,980, and the sample size is greater than 10,000, so we set the interval size to 10,000 and partition it into the interval [80,000, 90,000). Otherwise, for the Mr-En (\textit{Marathi-to-English}) language pair, the sample size is 5,080, which falls below 10,000. In this case, the interval size is defined as 5,000 and the Mr-En pair is categorized into the interval [5,000, 10,000).

In the second step, we calculate the sample mean for all language pairs within each sample interval. The sample mean serves as an approximation of the sample size for each language pair in the interval. While the sample mean cannot substitute the actual sample size of each language pair in the interval, we narrow down the interval size to minimize the disparity between the sample size of each language pair and the sample mean. Then, the sample intervals will be sorted in descending order based on sample mean values.
\begin{algorithm}[htb]  
  \caption{Framework of the incremental pre-training algorithm.}  
  \label{alg:Framwork}  
  \begin{algorithmic}[1]  
    \Require  
      The set of sample intervals sorted in descending order, $[S_1, S_2,..., S_n]$;
    \For {$S_i$ in $[S_1, S_2,..., S_n]$} 
        \State Set the sample mean in $S_{i}$, $M_{S_i}$
        \State Set the mean value of the untrained samples in $S_i$, $M_{ut}$
        \State Initialize $M_{ut} \gets M_{S_i}$
        \State Set the sample mean in $S_{i+1}$, $M_{S_{i+1}}$
        \While {$M_{ut}$ > $M_{S_{i+1}}$}
            \State Pre-train \textbf{BigTranslate} on $S_i$;
            \State Calculate $M_{ut}$ after each step of Pre-training;
        \EndWhile
        \State Add the untrained samples in $S_i$ to $S_{i+1}$
        \State Shuffle and Update $S_{i+1}$
    \EndFor
  \end{algorithmic}  
\end{algorithm}  

Finally, the model is exposed to all sample intervals following \textbf{Algorithm \ref{alg:Framwork}}. Initially, the model is exposed to the current sample interval, and dynamically calculates the mean value of the untrained samples in the current sample interval during training process.  When the mean value of the untrained sample is not greater than the sample mean of the next interval, we mix the current untrained sample and the next interval sample to form a new sample interval, and train model on the new sample interval. This process is repeated iteratively.

Assuming that the sample mean value of the sample interval $S_1$ is 103,251, denoted as $M_{S_1}$, and the sample mean value of the sample interval $S_2$ is 95,280, denoted as $M_{S_2}$. According to \textbf{Algorithm \ref{alg:Framwork}}, BigTranslate will be exposed to all samples in the $S_1$ interval for batch training. Following each training step, we will calculate the mean of the untrained samples in $S_1$, denoted as $M_{ut}$. With the increase of training steps, $M_{ut}$ will gradually approach $M_{S_2}$. When $M_{ut}$ is not greater than $M_{S_2}$ for the first time, we will merge the untrained samples in $S_1$ into $S_2$, and then start training on new $S_2$.

By adopting this incremental multilingual pre-training approach, we aim to ensure a more balanced and comprehensive model learned across different language pairs, thus addressing the challenge of uneven resource allocation during training. As a result, the model will successfully achieve mastery in 102 languages through its multilingual learning journey. We name the multilingually pre-trained model \textbf{BigTranslate}.

\subsection{Multilingual Translation Instruction Tuning}
Previous work \citep{mishra-etal-2022-cross, wei2022finetuned, sanh2022multitask} has shown that utilizing instruction-following data to tune LLMs enables such models to understand tasks described in natural languages, and show better multi-task learning ability on training tasks and generalization ability on unseen tasks. Instruction tuning does not inject new capabilities into the model. Instead, its purpose lies in activating the existing capabilities that were established during the training phase.

In this section, we construct a multilingual translation instruction dataset containing 102 languages and 242 language-pairs. Furthermore, we fine-tune the \textbf{BigTranslate} model to unlock the performance of model on multilingual machine translation ability.

\textbf{Instruction tuning data construction} The multilingual translation instruction tuning dataset consists of 1,000 parallel sentence pairs for each language pair, which is selected from the training set. If the number of sentence pairs in the training set is below 1,000, all training examples are chosen for the instruction fine-tuning data. Utilizing this approach, the instruction tuning dataset comprises 241,128 parallel sentence pairs across 242 language pairs.

\textbf{Instruction tuning prompts selection} We have designed a set of 28 multilingual translation prompts that encompass various application scenarios for multilingual translation. We randomly select a prompt from the set for instruction tuning for each parallel sentence. Accordingly, the instruction tuning dataset is scrambled to ensure randomness and diversity.

\subsection{Training Details}

The BigTranslate model is pre-trained on 42.8B and 47.0B tokens in the Chinese and Multilingual pre-training stages, respectively. 
The learning rates are empirically set to 5e-5 under a cosine scheduler with a 3\% warm-up percentage, and batch sizes are 65,536 in all pre-training stages.

Then, we fine-tune BigTranslate on 240k multilingual translation instructions, where the batch size and the number of epochs are set to 32 and 3, respectively. The learning rate and weight decay are empirically set to 2e-5 and 0. In inference, we employ beam search for decoding with a beam size of 5.

To speed up training, we adopted DeepSpeed \citep{rasley2020deepspeed} for data parallelization, FlashAttention \citep{tri2022flashattention}, and Gradient Checkpointing \citep{chen2016training} for saving memory.
All training processes are conducted on 16 A100 GPUs with 80GB of RAM.

\begin{figure*}[htbp]
    \centerline{
    \includegraphics[width=16cm]{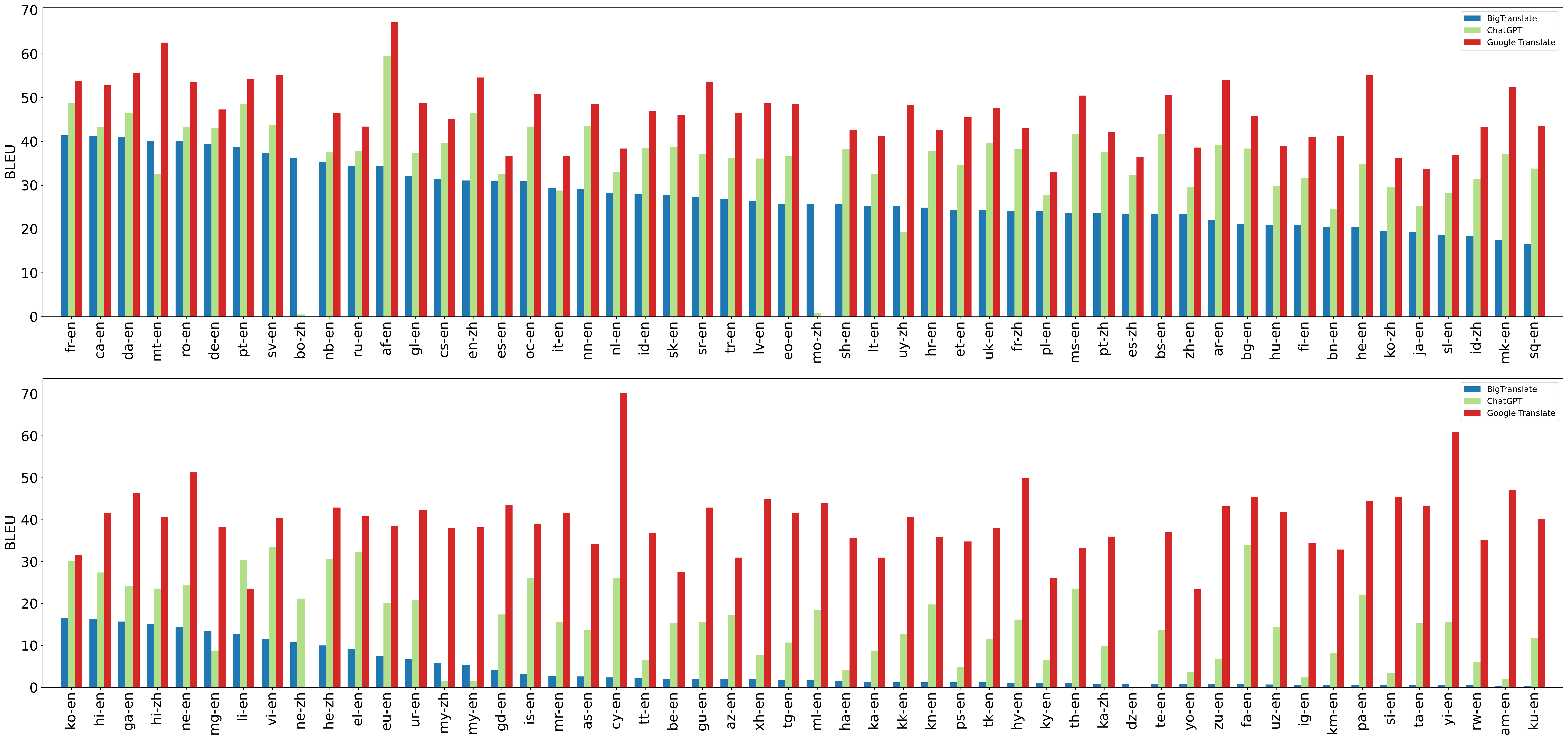}
    }
    \caption{An illustrated comparison of 102 languages from X to English or Chinese between BigTranslate, ChatGPT, and Google Translate. We sort the language scores in BLEU for BigTranslate in descending order.}
    \label{fig_BLEU}
\end{figure*}

\begin{figure*}[htbp]
    \centerline{
    \includegraphics[width=16cm]{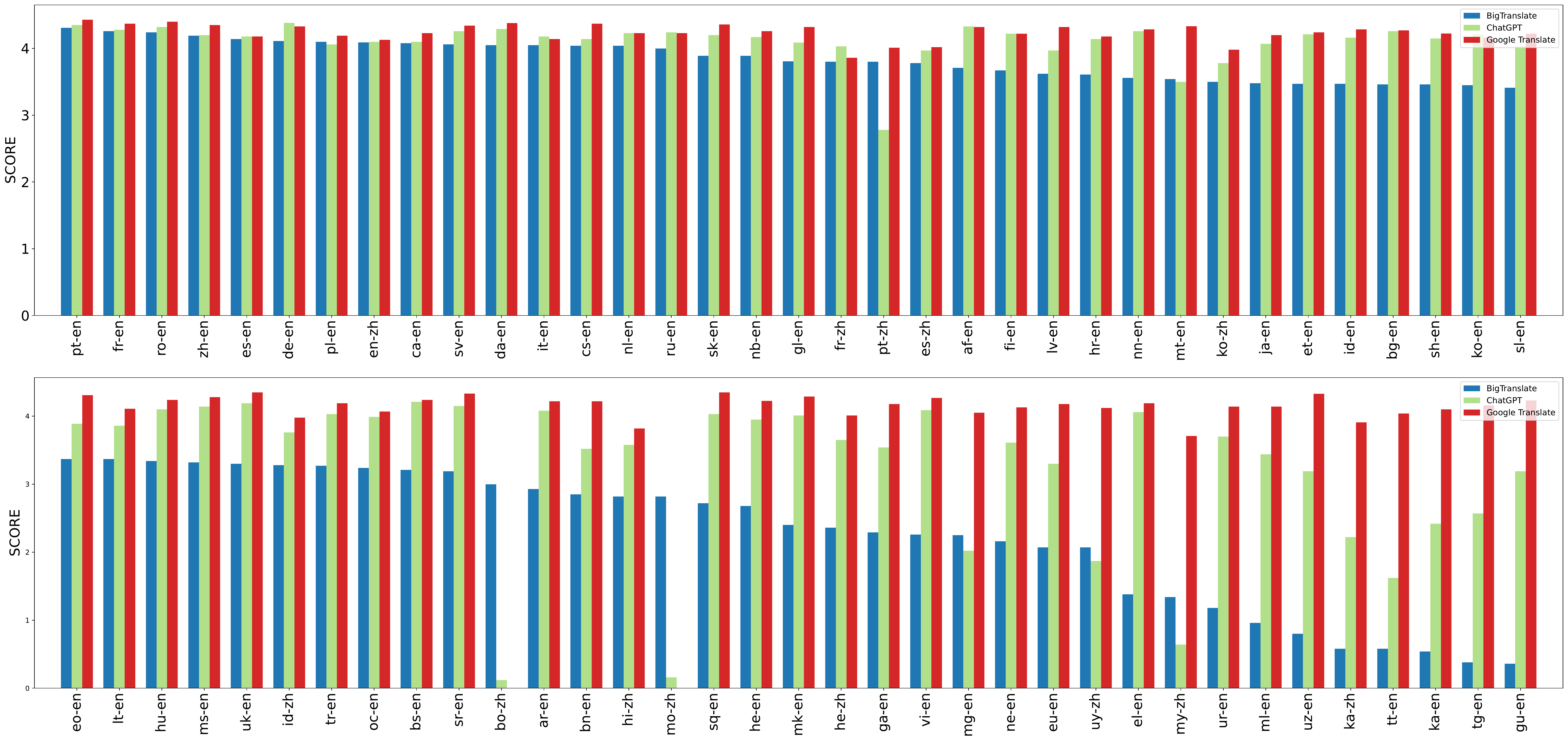}
    }
    \caption{An illustrated comparison of 70 languages from X to English or Chinese between BigTranslate, ChatGPT, and Google Translate. We sort the language scores in GPT-4 score for BigTranslate in descending order.}
    \label{fig_GPT}
\end{figure*}

\section{Experiments}
To demonstrate the effectiveness of the BigTranslate model, we conduct preliminary multilingual translation experiments on all 102 languages. We compare our model with Google Translate and ChatGPT\footnote{We use gpt-3.5-turbo API in May 2023}. We conduct the translation experiments from 102 languages to English or Chinese with the three systems\footnote{Since we plan to perform human-like evaluation and we cannot well evaluate the languages except English and Chinese, we just evaluate the translation direction from other languages to English or Chinese.}.

\subsection{Datasets and Evaluation Metrics} 
To assess the effectiveness of the multilingual machine translation model, we conducted evaluations using the devtest subset of the FLORES-200 dataset\footnote{We select 50 sentences in each direction of the devtest set for evaluation} \cite{costa2022no}. The FLORES-200 dataset comprises a corpus extracted from 842 web articles encompassing various fields and topics, resulting in a total of 3,001 sentences. Notably, a team of experts meticulously translated these sentences into 200 languages, thereby constituting the FLORES-200 dataset.

\subsection{Main Results}
In this section, we report the main results on multilingual machine translation on BigTranslate, ChatGPT\footnote{gpt-3.5-turbo API}, and Google Translate. Then, we report our main findings about the exploration of multilingual machine translation using a large language model.

\textbf{Automatic Evaluation with BLEU}~~Figure \ref{fig_BLEU} shows the detailed outcomes in BLEU scores of the translation results from 102 languages into English and Chinese. Detailed results for each translation direction are listed in Appendix \ref{appedix_b}. In Figure \ref{fig_BLEU}, we have sorted the BLEU scores obtained from the BigTranslate model in descending order. We split the whole figure into two parts for clarity. The upper figure presents the first half of all the language pairs, while the bottom figure presents the second half. Notably, the upper figure reveals that a significant proportion of language pairs in the BigTranslate model, specifically 46 out of 104 (equivalent to over 44\%), yield more than 20 BLEU scores. This observation suggests that the BigTranslate model exhibits commendable performance with respect to these languages, if we believe BLEU is a reliable evaluation metric. We also find that, when comparing our BigTranslate model with ChatGPT, the results demonstrate that BigTranslate performs on par with ChatGPT in many languages. Intriguingly, BigTranslate surpasses ChatGPT in 9 language pairs (e.g. mt-en, bo-zh, it-en, mo-zh, uy-zh, mg-en, my-zh, my-en, dz-en) in terms of BLEU scores. When comparing BigTranslate with Google Translate, Figure \ref{fig_BLEU} clearly shows that BigTranslate significantly lags behind Google Translate in most of the language pairs in BLEU scores. However, as \cite{hendy2023good, anil2023palm} pointed out that BLEU is not a good evaluator when the BLEU scores exceed a threshold (e.g. 20.0), and it has a low correlation with human preference.

\textbf{Automatic Evaluation with GPT-4}~~\cite{liu2023geval} demonstrates that GPT-4 achieves a much higher Spearman correlation with human than all previous methods on the summarization task. We thus follow \citep{liu2023geval}, and employ GPT-4 with Chain-of-Thoughts (CoT) prompts to automatically evaluate the quality of translation. In the evaluation, GPT-4 assigns a score ranging from 0 to 5 to assess the translation quality of every sentence across three models. The prompt is a natural language instruction and contains two parts. In the first part, we define the evaluation task and explain the scoring scope. The first part of the prompt we use is below.

\begin{quote}
\itshape
You will be given two sentences, translated sentence is translated from source sentence, reference sentence is the ground truth of translation.

Your task is to rate the translation result between translated sentence and reference sentence.  

Assign a score for translation result on a scale of 0 to 5, where 0 is the lowest and 5 is the highest based on the Evaluation Criteria. 
\end{quote}

In the second part, the prompt should contain specific evaluation criteria for evaluating machine translation, which can guide GPT-4 on how to score. We describe evaluation criteria below:

\begin{quote}
\itshape
Semantic similarity refers to the measurement of how similar or related two sentences are in terms of their meaning or semantics. It focuses on capturing the similarity in the underlying concepts, ideas, or information conveyed by the sentences, rather than just the surface-level lexical or syntactic similarities.

The translated sentence can completely express the meaning of the reference sentence. The closer the translated sentence is to the reference sentence, the higher the score.

The style of the translated sentence should be as consistent as possible with the reference sentence
\end{quote}

Subsequently, we calculate the average score of all sentences within the same language pairs, which represents the overall translation score of the respective language pair. In Appendix \ref{appedix_a}, we can see the details of the GPT-4 evaluation prompt.

We utilize GPT-4 to rate the translation of 70 language pairs for three translation models. The results are illustrated in Figure \ref{fig_GPT} and the detailed results can be found in Appendix \ref{appedix_c}. The figure indicates that a total of 28 language pairs exhibit good or excellent translation scores surpassing 3.5, thereby accounting for 40\% of all language pairs. This result demonstrates that BigTranslate has remarkable multilingual translation performance under GPT-4 evaluation. If we compare Figure \ref{fig_GPT} against Figure \ref{fig_BLEU}, we can observe some interesting phenomena. For example, the performance gap between our BigTranslate and Google Translate becomes much narrowed in many language pairs, indicating that BigTranslate is approaching Google Translate in dozens of languages in terms of GPT-4 score. In comparison to ChatGPT, we can find that BigTranslate achieves similar performance to ChatGPT in 27 languages, with a difference in GPT-4 scores of less than 0.3 points. Moreover, BigTranslate outperforms ChatGPT in 8 languages in GPT-4 scores.

\subsection{Discussion}
The experimental results displayed in the previous section demonstrate the effectiveness of our BigTranslate model in extending large language models to enable multilingual translation over 100 languages.
As a multilingual translator, BigTranslate can be employed in translation for many languages, although we still need to further improve the translation quality for extremely low-resource languages.
In addition to language translation, BigTranslate can also be applied in other natural language processing tasks just as ChatGPT does. Notably, the LLM ability especially the English capability can be transferred to many other languages including several low-resource ones with the help of our BigTranslate model. For example, BigTranslate is now good at translating Tibetan and Mongolian languages, and English and Chinese NLP abilities in LLMs can be transferred into these languages.

\section{Conclusion and Future Work}
In this work, we introduced BigTranslate which is a large language model equipped with the capability of multilingual translation over 100 natural languages. After two steps of continuing training with massive Chinese monolingual data and large scale multilingual parallel data of 102 languages, LLaMA is extended to have the potential multilingual ability on 102 natural languages. Using the final step of instruction tuning with multilingual translation data, BigTranslate is obtained. The experiments demonstrate that BigTranslate performs comparable to Google Translate and ChatGPT in many languages, and even surpasses ChatGPT in 8 languages when evaluated with GPT-4.

Due to the issue of unbalanced data, BigTranslate is still weak in dozens of languages. In the future, we will extend BigTranslate to further enhance its ability in low-resource languages. Moreover, we will take full advantage of the multilingual ability of BigTranslate, and improve the performance of the languages in other natural language processing tasks.


\section*{Acknowledgements}
We thank Jinliang Lu, Yu Lu, Yangyifan Xu and Qian Wang for multilingual translation data construction.

\bibliography{anthology,custom}
\bibliographystyle{acl_natbib}

\clearpage
\onecolumn
\appendix
\section{GPT-4 prompt for Evaluating Machine Translation}
\label{appedix_a}
\textbf{Example Prompt:}
\begin{quote}
\itshape
You will be given two sentences, translated sentence is translated from source sentence, reference sentence is the ground truth of translation.

Your task is to rate the translation result between translated sentence and reference sentence.  

Assign a score for translation result on a scale of 0 to 5, where 0 is the lowest and 5 is the highest based on the Evaluation Criteria. 
\\
\\
Evaluation Criteria: 

Semantic similarity refers to the measurement of how similar or related two sentences are in terms of their meaning or semantics. It focuses on capturing the similarity in the underlying concepts, ideas, or information conveyed by the sentences, rather than just the surface-level lexical or syntactic similarities.

The translated sentence can completely express the meaning of the reference sentence. The closer the translated sentence is to the reference sentence, the higher the score.

The style of the translated sentence should be as consistent as possible with the reference sentence
\\
\\
Sample 1:

Translated Sentence: \{\} 

Reference Sentence: \{\} 

...

Sample 5:

Translated Sentence: \{\} 

Reference Sentence: \{\}

Evaluation Form (Please output score ONLY):

-Overall rating
\end{quote}

\section{Detailed Results on 102 Languages with BLEU}
Detailed results of our evaluated models on 102 languages with BLEU are shown in Table \ref{table1}.
\label{appedix_b}
\begin{table}[htbp]
    \centering
    \caption{Detailed results on 102 languages with BLEU}
    \resizebox{0.95\linewidth}{!}{
    \begin{tabular}{cccc|cccc}
        \toprule
        Language pair & BigTranslate & ChatGPT & Google Translate & Language pair & BigTranslate & ChatGPT & Google Translate \\
        \midrule
        fr-en&41.4&48.8&53.8&ko-en&16.5&30.2&31.6\\
        ca-en&41.2&43.3&52.8&hi-en&16.3&27.4&41.6\\
        da-en&41.0&46.4&55.6&ga-en&15.7&24.2&46.3\\
        mt-en&40.1&32.5&62.6&hi-zh&15.1&23.6&40.7\\
        ro-en&40.1&43.3&53.5&ne-en&14.4&24.5&51.3\\
        de-en&39.5&43.0&47.3&mg-en&13.5&8.8&38.3\\
        pt-en&38.7&48.6&54.2&li-en&12.7&30.3&23.5\\
        sv-en&37.3&43.8&55.2&vi-en&11.6&33.4&40.5\\
        bo-zh&36.3&0.5&0.0&ne-zh&10.8&21.2&0.0\\
        nb-en&35.4&37.5&46.4&he-zh&10.0&30.6&42.9\\
        ru-en&34.5&37.9&43.4&el-en&9.2&32.3&40.8\\
        af-en&34.4&59.5&67.2&eu-en&7.5&20.1&38.6\\
        gl-en&32.1&37.4&48.8&ur-en&6.7&20.9&42.4\\
        cs-en&31.4&39.6&45.2&my-zh&5.9&1.6&38.0\\
        en-zh&31.1&46.6&54.6&my-en&5.3&1.5&38.2\\
        es-en&30.9&32.6&36.7&gd-en&4.1&17.4&43.6\\
        oc-en&30.9&43.4&50.8&is-en&3.2&26.1&38.9\\
        it-en&29.4&28.8&36.7&mr-en&2.8&15.6&41.6\\
        nn-en&29.2&43.5&48.6&as-en&2.6&13.6&34.2\\
        nl-en&28.2&33.1&38.4&cy-en&2.4&26.0&70.2\\
        id-en&28.1&38.5&46.9&tt-en&2.3&6.5&36.9\\
        sk-en&27.8&38.8&46.0&be-en&2.1&15.4&27.5\\
        sr-en&27.4&37.1&53.5&gu-en&2.0&15.6&42.9\\
        tr-en&26.9&36.3&46.5&az-en&2.0&17.3&31.0\\
        lv-en&26.4&36.1&48.7&xh-en&1.9&7.8&44.9\\
        eo-en&25.8&36.6&48.5&tg-en&1.8&10.7&41.6\\
        mo-zh&25.7&0.9&0.0&ml-en&1.7&18.5&44.0\\
        sh-en&25.7&38.3&42.6&ha-en&1.5&4.2&35.6\\
        lt-en&25.2&32.6&41.3&ka-en&1.3&8.6&31.0\\
        uy-zh&25.2&19.3&48.4&kk-en&1.2&12.8&40.6\\
        hr-en&24.9&37.8&42.6&kn-en&1.2&19.8&35.9\\
        et-en&24.4&34.6&45.5&ps-en&1.2&4.8&34.8\\
        uk-en&24.4&39.7&47.6&tk-en&1.2&11.5&38.1\\
        fr-zh&24.2&38.2&43.0&hy-en&1.1&16.2&49.9\\
        pl-en&24.2&27.8&33.0&ky-en&1.1&6.6&26.1\\
        ms-en&23.7&41.6&50.5&th-en&1.1&23.6&33.2\\
        pt-zh&23.6&37.6&42.2&ka-zh&0.9&9.9&36.0\\
        es-zh&23.5&32.3&36.4&dz-en&0.9&0.2&0.0\\
        bs-en&23.5&41.6&50.6&te-en&0.9&13.7&37.1\\
        zh-en&23.4&29.6&38.6&yo-en&0.9&3.7&23.4\\
        ar-en&22.1&39.1&54.1&zu-en&0.9&6.8&43.2\\
        bg-en&21.2&38.4&45.8&fa-en&0.8&34.0&45.4\\
        hu-en&21.0&29.9&39.0&uz-en&0.7&14.3&41.9\\
        fi-en&20.9&31.6&41.0&ig-en&0.6&2.4&34.5\\
        bn-en&20.5&24.6&41.3&km-en&0.6&8.3&32.9\\
        he-en&20.5&34.8&55.1&pa-en&0.6&22.0&44.5\\
        ko-zh&19.6&29.6&36.3&si-en&0.6&3.4&45.5\\
        ja-en&19.4&25.3&33.7&ta-en&0.6&15.3&43.4\\
        sl-en&18.6&28.2&37.0&yi-en&0.6&15.6&60.9\\
        id-zh&18.4&31.5&43.3&rw-en&0.5&6.1&35.2\\
        mk-en&17.5&37.2&52.5&am-en&0.3&2.0&47.1\\
        sq-en&16.6&33.8&43.5&ku-en&0.3&11.8&40.2\\
        \bottomrule
    \end{tabular}
    \label{table1}
    }
\end{table}

\section{Detailed Results on 70 Languages with GPT-4 Evaluation}
\label{appedix_c}
Detailed results of our evaluated models on 70 languages with GPT-4 score are shown in Table \ref{table2}.
\begin{table}[htbp]
    \centering
    \caption{Detailed results on 70 languages with GPT-4 Evaluation}
    \resizebox{0.95\linewidth}{!}{
    \begin{tabular}{cccc|cccc}
        \toprule
        Language pair & BigTranslate & ChatGPT & Google Translate & Language pair & BigTranslate & ChatGPT & Google Translate \\
        \midrule
        pt-en&4.31&4.35&4.43&eo-en&3.37&3.89&4.31\\
        fr-en&4.26&4.28&4.37&lt-en&3.37&3.86&4.11\\
        ro-en&4.24&4.32&4.4&hu-en&3.34&4.1&4.24\\
        zh-en&4.19&4.2&4.35&ms-en&3.32&4.14&4.28\\
        es-en&4.14&4.18&4.18&uk-en&3.3&4.19&4.35\\
        de-en&4.11&4.384&4.33&id-zh&3.28&3.76&3.98\\
        pl-en&4.1&4.06&4.19&tr-en&3.27&4.03&4.19\\
        en-zh&4.09&4.1&4.13&oc-en&3.24&3.99&4.07\\
        ca-en&4.08&4.1&4.23&bs-en&3.21&4.21&4.24\\
        sv-en&4.06&4.26&4.34&sr-en&3.19&4.15&4.334\\
        da-en&4.05&4.29&4.38&bo-zh&3.0&0.12&0.0\\
        it-en&4.05&4.18&4.14&ar-en&2.93&4.08&4.22\\
        cs-en&4.04&4.14&4.37&bn-en&2.85&3.52&4.22\\
        nl-en&4.04&4.23&4.23&hi-zh&2.82&3.58&3.82\\
        ru-en&4.0&4.24&4.23&mo-zh&2.82&0.16&0.0\\
        sk-en&3.89&4.2&4.36&sq-en&2.72&4.03&4.35\\
        nb-en&3.89&4.17&4.26&he-en&2.68&3.95&4.226\\
        gl-en&3.808&4.088&4.32&mk-en&2.4&4.01&4.29\\
        fr-zh&3.8&4.03&3.86&he-zh&2.36&3.65&4.01\\
        pt-zh&3.8&2.78&4.01&ga-en&2.29&3.54&4.18\\
        es-zh&3.78&3.97&4.02&vi-en&2.26&4.09&4.27\\
        af-en&3.71&4.33&4.32&mg-en&2.25&2.02&4.05\\
        fi-en&3.67&4.22&4.22&ne-en&2.16&3.61&4.13\\
        lv-en&3.62&3.97&4.32&eu-en&2.07&3.3&4.18\\
        hr-en&3.61&4.14&4.18&uy-zh&2.07&1.87&4.12\\
        nn-en&3.56&4.26&4.284&el-en&1.38&4.06&4.19\\
        mt-en&3.54&3.5&4.332&my-zh&1.34&0.64&3.71\\
        ko-zh&3.5&3.78&3.98&ur-en&1.18&3.7&4.14\\
        ja-en&3.48&4.07&4.2&ml-en&0.96&3.44&4.14\\
        et-en&3.47&4.21&4.24&uz-en&0.8&3.19&4.33\\
        id-en&3.47&4.16&4.286&ka-zh&0.58&2.22&3.91\\
        bg-en&3.46&4.26&4.27&tt-en&0.58&1.62&4.04\\
        sh-en&3.46&4.15&4.222&ka-en&0.54&2.42&4.1\\
        ko-en&3.45&4.21&4.16&tg-en&0.38&2.57&4.16\\
        sl-en&3.41&4.03&4.22&gu-en&0.36&3.19&4.23\\
        \bottomrule
    \end{tabular}
    \label{table2}
    }
\end{table}

\section{The corresponding table of the language code and its name}
Involved languages and the corresponding language codes in this paper are listed in Table 
 \ref{table3}.
\label{appedix_d}
\begin{table}[htbp]
    \centering
    \caption{The corresponding table of the language code and its name}
    \resizebox{0.95\linewidth}{!}{
    \begin{tabular}{cc|cc|cc}
        \toprule
        Language code  & Language & Language code  & Language & Language code  & Language\\
        \midrule
        af&Afrikaans&he&Hebrew&or&Oriya\\
        am&Amharic&hi&Hindi&pa&Panjabi\\
        an&Aragonese&hr&Croatian&pl&Polish\\
        ar&Arabic&hu&Hungarian&ps&Pashto\\
        as&Assamese&hy&Armenian&pt&Portuguese\\
        ast&Asturian&id&Indonesian&ro&Romanian\\
        az&Azerbaijani&ig&Igbo&ru&Russian\\
        be&Belarusian&is&Icelandic&rw&Kinyarwanda\\
        bg&Bulgarian&it&Italian&se&Northern Sami\\
        bn&Bengali&ja&Japanese&sh&Serbo-Croatian\\
        bo&Tibetan&ka&Georgian&si&Sinhala\\
        br&Breton&kk&Kazakh&sk&Slovak\\
        bs&Bosnian&km&Central Khmer&sl&Slovenian\\
        ca&Catalan&kn&Kannada&sq&Albanian\\
        cs&Czech&ko&Korean&sr&Serbian\\
        cy&Welsh&ku&Kurdish&sv&Swedish\\
        da&Danish&ky&Kyrgyz&ta&Tamil\\
        de&German&li&Limburgan&te&Telugu\\
        dz&Dzongkha&lt&Lithuanian&tg&Tajik\\
        el&Greek&lv&Latvian&th&Thai\\
        en&Engilish&mg&Malagasy&tk&Turkmen\\
        eo&Esperanto&mk&Macedonian&tr&Turkish\\
        es&Spanish&ml&Malayalam&tt&Tatar\\
        et&Estonian&mo&Mongolian&uk&Ukrainian\\
        eu&Basque&mr&Marathi&ur&Urdu\\
        fa&Persian&ms&Malay&uy&Uighur\\
        fi&Finnish&mt&Maltese&uz&Uzbek\\
        fr&French&my&Burmese&vi&Vietnamese\\
        fy&Western Frisian&nb&Norwegian Bokmal&wa&Walloon\\
        ga&Irish&ne&Nepali&xh&Xhosa\\
        gd&Gaelic&nl&Dutch&yi&Yiddish\\
        gl&Galician&nn&Norwegian Nynorsk&yo&Yoruba\\
        gu&Gujarati&no&Norwegian&zh&Chinese\\
        ha&Hausa&oc&Occitan&zu&Zulu\\
        \bottomrule
    \end{tabular}
    \label{table3}
    }
\end{table}

\end{CJK*}
\end{document}